\documentclass{article} 
\usepackage{iclr2024_conference,times}

\usepackage{amsmath,amsfonts,bm}

\def\eqref#1{equation~\ref{#1}}

\def\1{\bm{1}}

\DeclareMathAlphabet{\mathsfit}{\encodingdefault}{\sfdefault}{m}{sl}
\SetMathAlphabet{\mathsfit}{bold}{\encodingdefault}{\sfdefault}{bx}{n}

\usepackage{subcaption}

\usepackage{algorithm}
\usepackage{multirow}
\usepackage[noend]{algorithmic}
\usepackage{wrapfig}
\usepackage{framed}
\usepackage{hyperref}
\usepackage{url}
\usepackage{times}
\usepackage{latexsym}
\usepackage{booktabs}
\usepackage{graphicx}
\usepackage{caption}
\usepackage{subcaption}
\usepackage{amsmath}
\usepackage{multirow}
	
\usepackage{color, colortbl}
\usepackage{pgfplots}
\usetikzlibrary{patterns}
\usepackage{filecontents}
\usepackage{algorithmic,algorithm}
\usepackage{csvsimple}
\usepackage{amsmath}

\definecolor{blue}{HTML}{4285f4}
\definecolor{LightGrey}{HTML}{d9d9d9}

\usepackage{microtype}
\newcommand{\model}{\textsc{\textbf{RECOMP}}~}

\newcommand{\modelnospace}{\textsc{\textbf{RECOMP}}}

\title{RECOMP: Improving Retrieval-Augmented LMs with Compression and Selective Augmentation}

\author{Fangyuan Xu$^1$, Weijia Shi$^2$, Eunsol Choi$^1$\\
Department of Computer Science\\
$^1$The University of Texas at Austin\\
$^2$University of Washington\\
\texttt{\{fangyuan,eunsol\}@utexas.edu} , \texttt{swj0419@cs.washington.edu} \\
}

\iclrfinalcopy 
\begin{document}

\maketitle

\begin{abstract}

Retrieving documents and prepending them in-context at inference time improves performance of language model (LMs) on a wide range of tasks. However, these documents, often spanning hundreds of words, make inference substantially more expensive. We propose compressing the retrieved documents into textual summaries prior to in-context integration. This not only reduces the computational costs but also relieves the burden of LMs to identify relevant information in long retrieved documents. We present two compressors -- an \textit{extractive compressor} which selects useful sentences from retrieved documents  and an \textit{abstractive compressor} which generates summaries by synthesizing information from multiple documents. 
Both compressors are trained to improve LMs' performance on end tasks when the generated summaries are prepended to the LMs' input, while keeping the summary concise.
If the retrieved documents are irrelevant to the input or offer no additional information to LM, our compressor can return an empty string, implementing selective augmentation.
We evaluate our approach on language modeling task and open domain question answering task. We achieve a compression rate of as low as 6\% with minimal loss in performance for both tasks, significantly outperforming the off-the-shelf summarization models. We show that our compressors trained for one LM can transfer to other LMs on the language modeling task and provide summaries largely faithful to the retrieved documents.\footnote{Our code is available at \url{https://github.com/carriex/recomp}.}
\end{abstract}

\section{Introduction}

Retrieval-augmented language models (RALMs) \citep{Khandelwal2019GeneralizationTM, Izacard2022FewshotLW, Lewis2020RetrievalAugmentedGF, pmlr-v162-borgeaud22a} have shown impressive performance on knowledge-intensive tasks \citep{kwiatkowski2019natural, petroni-etal-2021-kilt}. Simply prepending retrieved documents to the input without updating the language models (LMs) ~\citep{Shi2023REPLUGRB,Ram2023InContextRL,Si2022PromptingGT} allows retrieval augmentation even for black-box LMs, but such approach comes with limitations. First, it increases computational costs as LMs now encode substantially more tokens. 
Second, even if we manage to adapt LMs to efficiently incorporate longer context \citep{Beltagy2020LongformerTL, Zaheer2020BigBT}, these models struggle to use all information in the context, frequently missing information placed in the middle~\citep{liu2023lost}. Third, prepending a large number of documents in-context can further \textit{confuse} LMs with irrelevant information, degrading model performances~\citep{Mallen2022WhenNT, Shi2023LargeLM}.

\begin{figure*}[t!]
 \vspace{-2em}
    \centering
    \includegraphics[trim={0 0 0 0},clip,width=\textwidth]{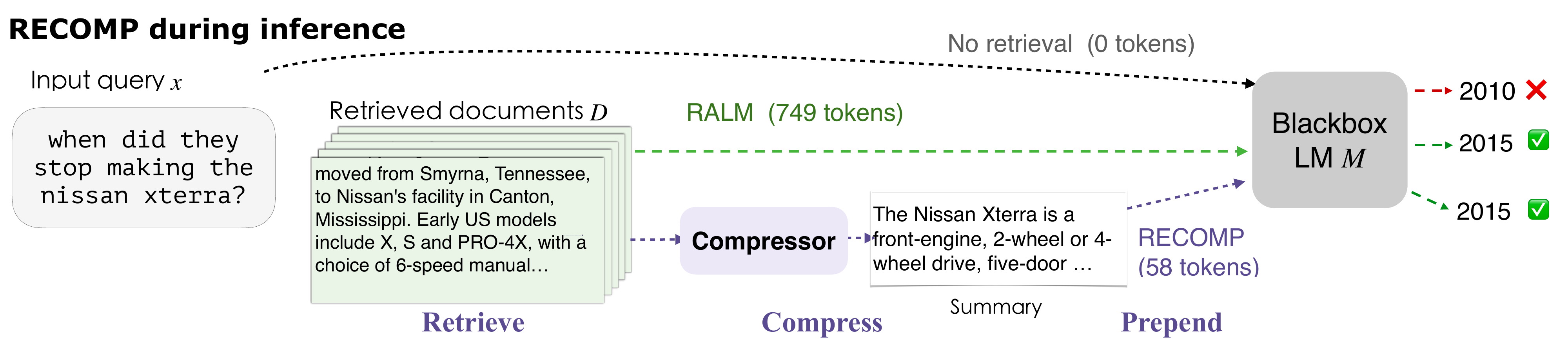}
    \caption{An illustration of \modelnospace, which compresses retrieved documents into a texual summary before prepending it as input to a language model at inference time. The compressed summary guides the LM to generate the correct answer, while significantly reducing the computation costs required to encode the documents.}
    \label{fig:intro}
\end{figure*}

To overcome such limitations, we propose \model  (\textbf{Re}trieve, \textbf{Com}press, \textbf{P}repend), an intermediate step for RALMs which compresses retrieved documents into a textual summary prior to in-context augmentation. Figure \ref{fig:intro} illustrates our approach. The generated summary should be concise to maximize efficiency, be faithful to the retrieved evidence documents, and guide RALM to generate desired outputs when prepended to the input. To satisfy both efficiency and effectiveness constraints, our compressor strategically performs selective augmentation by generating an empty summary when the retrieved documents are irrelevant or unhelpful for target task.

We propose compressors: (1) \textit{Extractive compressor} which selects relevant sentences from retrieved document set; (2) \textit{Abstractive compressor} which generates a summary synthesizing information from multiple retrieved documents. Both compressors implement multi-document query-focused summarization~\citep{Xu2020CoarsetoFineQF}, where we summarize retrieved evidence document set with respect to the input query. As we aim to enable RALM to generate correct output when summary is prepended to the input query, we design training schemes to optimize the end task performance. Our extractive compressor is trained with a contrastive learning objective to identify sentences that lead to target outputs, and our abstractive compressor is distilled ~\citep{west-etal-2022-symbolic} from an extreme-scale LM (e.g. GPT-3), which achieves impressive summarization performance.

Our experiments show that \model can improve performance of frozen LMs on language modeling \citep{Merity2016PointerSM} and three question answering datasets (Natural Questions \citep{kwiatkowski2019natural}, TriviaQA \citep{triviaqa} and HotpotQA \citep{yang2018hotpotqa}), while prepending significantly fewer tokens compared to RALM without compression.  {We present two oracle compression methods -- an extractive oracle which selects a sentence in evidence documents that leads to the best task performance and an abstractive oracle which chooses between a summary generated by extreme-scale LLM (e.g. GPT-3) and no retrieval augmentation that leads to the best task performance.} Both oracle methods achieve a compression rate as low as 6\% and significantly \textit{outperforms} prepending full documents. Our trained compressors also show promising results. For language modelling, both trained compressors achieve a compression ratio of 25\% with minimal performance drop. When applied to QA datasets, our best model compresses the documents to 5 - 10\% of the original tokens with at most less than 10\% relative performance drop. We conclude with careful analyses of our approach that reveal both its strength and weaknesses, thereby building foundation for future work.

\section{Problem Formulation: \model}

Given an input sequence $\mathbf{x}$, a target output sequence $\mathbf{y}$ and a set of $N$ retrieved documents  $D$ ($\left[  d_1 , d_2,  ...d_N \right]$),\footnote{Improving retriever is not the focus of this work, so we assume a set of retrieved documents are provided.} \model compresses retrieved documents $D$ with respect to $\mathbf{x}$ into a summary $\mathbf{s}$ which captures core information in $D$ relevant to $\mathbf{x}$ with significantly fewer tokens than $D$. Our architecture consists of two modules: compressor $c_\theta$ and LM $M$. In this work, we assume a blackbox LM and train the compressor. Given the set of retrieved $N$ documents ($\left[  d_1 , d_2,  ...d_N \right]$) and the input sequence $\mathbf{x}$, a compressor returns a token sequence $\mathbf{s}$. We design our compressor to be substantially smaller than LM $M$, as we aim to reduce computational costs of encoding a set of retrieved documents. 

The output from compressor, $\mathbf{s}$, should be: (1) \textbf{Concise}: The summary should be as short as possible to optimize efficiency. If the retrieved documents do not contain relevant information or retrieval augmentation is not necessary, $\mathbf{s}$ can be an empty sequence. {(2) \textbf{Effecive}: }when $\mathbf{s}$ is prepended to input sequence $\mathbf{x}$ and provided to LM $M$ as a prompt, LM should generate the target output sequence $\mathbf{y}$. {(3) \textbf{Faithful}: }$\mathbf{s}$ should be a faithful and interpretable summary of the input document set (i.e., $\mathbf{s}$ must be entailed by the input document set ($\left[  d_1, d_2, ...d_N \right]$)). We focus on training compressors for conciseness and effectiveness. We summarize the key ideas for our two compressors, {extractive compressors} and {abstractive compressor} here, and discuss their training schemes formally in Section~\ref{sec:compressor}.

\paragraph{Extractive Compressor} Given $n$ sentences $\left[\mathbf{s_{1}, s_{2} ... s_{n}}\right]$ in the input document set ($\left[  d_1, d_2, ...d_N \right]$), we train a dual encoder model $\mathtt{enc}_\theta$ which embeds sentence $\mathbf{s_i}$ and the input sequence $\mathbf{x}$ into fixed-dimensional embeddings respectively. Their inner product represents how helpful it would be for the LM $M$ to prepend $\mathbf{s_i}$ to the input $\mathbf{x}$ to generate $\mathbf{y}$. The final summary $\mathbf{s}$ from the compressor will be a concatenation of top N sentences ranked by their inner product with the input. As this approach is extractive, we assume the faithfulness criteria is mostly satisfied.\footnote{Recent work~\citep{Zhang2022ExtractiveIN} shows that extractive approach does not always preserve faithfulness, but such cases are still rare compared to abstractive approaches which can easily hallucinate.}

\paragraph{Abstractive Compressor} We train an encoder-decoder model $\mathtt{encdec}_\theta$ to serve as an abstractive compressor, which takes the input sequence $\mathbf{x}$ and a concatenation of retrieved document set $D$ $\left[  d_1 ; d_2; ...d_N \right]$) and output a summary $\mathbf{s}$. Although we do not have human annotations to train this model, prior work~\citep{goyalzeroshotnews2022,Chen2023ComplexCV,potluri-etal-2023-concise} suggests that the extreme-scale LMs can generate good query-focused summaries when prompted carefully. Yet, using an extreme-scale model as the compressor is not desirable as we want the compressor to be substantially smaller than the LMs. Thus, we perform \textbf{distillation}~\citep{Hinton2015DistillingTK} of extreme-scale LMs to build a lightweight abstractive compressor $\mathtt{encdec}_\theta$. 
We do not train specifically for faithfulness, but later manually evaluate the faithfulness in Section~\ref{sec:analysis}.

\vspace{-0.3em}
\section{Learning the compressors}\label{sec:compressor}

\vspace{-0.2em}
{Our compressor resembles text summarization models in the output should be faithful to the original input, yet the main goal is different. Instead of capturing salient information for humans readers, compressors aim to produce a concise text that are useful for a LM on an end task.} In this section, we describe how to train the extractive compressor (\S \ref{section:extractive}) and the abstractive compressor (\S \ref{section:abstractive}) {leveraging end task signals}. Further training details can be found in the Appendix \ref{sec:training_detail}. 
\vspace{-0.3em}
\subsection{Extractive Compression} \label{section:extractive}

As we formulate extractive compression as a ranking problem, training extractive compressor resembles training a reranker for the retrieved documents\footnote{\citet{Ram2023InContextRL} proposes a document reranker based on a cross-encoder model, which is a similar set-up to our sentence selector, but less compute efficient.} with two differences. First, our compressor considers a different granularity of input (sentence) compared to the initial retrieval unit (paragraph). Second, the sentence is evaluated based on whether it is useful as input for the LM $M$ on the downstream task~\citep{Shi2023REPLUGRB,Ram2023InContextRL}.

\begin{wrapfigure}{r}{0.5585\textwidth}
\vspace{-2.4em}
\begin{framed}
\small
\textbf{Input:} Base LM $M$, Compressor $\mathtt{enc}_\theta$, 
Training data $\{\mathbf{x_i, S_i, y_i}\}_1^T$ where $\mathbf{x_i}$ is input, $\mathbf{S_i}=\{\mathbf{s_j}\}_1^n$ is a set of candidate sentences from the retrieved documents for $\mathbf{x_i}$, $\mathbf{y_i}$ is the target answer, and score threshold  $\epsilon$.

\textbf{Output:} An updated extractive compressor encoder $\texttt{enc}_\theta$


\begin{algorithmic}[1]
\STATE $\mathcal{T} \gets \emptyset$

\FOR{$i  \in \{1, \dots, T \}$} 
    \STATE $\mathbf{p_{i}} \gets \mathrm{argMax}_{\mathbf{s_j} \in \{\mathbf{S_i}\}} \textbf{Score}(M, \mathbf{y_i}, \left[  \mathbf{s_j}; \mathbf{x_i}  \right])$ \label{alg:positive}

    \FOR{$j  \in \{1, \dots, n \}$} 
     \STATE $\mathcal{L} \gets \emptyset $
        \IF{$\textbf{Score}(M, \mathbf{y_i}, \left[  \mathbf{s_j}; \mathbf{x_i}  \right]) + \epsilon < \textbf{Score}(M, \mathbf{y_i}, \left[  \mathbf{p_i}; \mathbf{x_i}  \right]) $}\label{alg:criteria}
            \STATE{$\mathcal{L} \gets \mathcal{L} \cup \mathbf{s_i}  $ }    
    \ENDIF
    \ENDFOR
    \IF{$|\mathcal{L}|>0$}
    \STATE $\mathcal{N}_i \gets \mathrm{argTop}5_{\mathbf{s_j} \in \mathcal{L}} (\langle \mathtt{enc}_\theta(\mathbf{s_j}), \mathtt{enc}_\theta(\mathbf{x_i}) \rangle) $

\STATE $\mathcal{T} \gets \mathcal{T} \cup \{ (\mathbf{x_i, p_i}, \mathcal{N}_i) \}$
\ENDIF
\ENDFOR



\STATE $\mathtt{enc}_\theta=\textbf{Finetune}(\mathtt{enc}_\theta, \mathcal{T})$

\end{algorithmic}
\end{framed}
\vspace{-0.8em}
\caption{Learning an extractive compressor for language modeling task. }\label{alg:contrastive}
\vspace{-2em}
\end{wrapfigure}

\paragraph{Model} We train a dual-encoder model $\mathtt{enc}_\theta$ which encodes the input context $x$ and the candidate sentence $\mathbf{s_{i}}$ separately. We obtain an embedding of $\mathbf{x}$ and $\mathbf{s_{i}}$ by taking the representation of the \texttt{[CLS]} token respectively, and compute their similarity by calculating the inner product of the two. We initialize our model with the contriever checkpoint \citep{izacard2021contriever}. This model consists of 110M parameters, satisfying the efficiency desideratum of compressor.  

\paragraph{Training} \autoref{alg:contrastive} presents pseudocode for training an extractive compressor with contrastive loss for the language modeling task. For each input query $\mathbf{x_i}$, we identify positive and negative sentences from retrieved documents.

For each pair of input sequence $\mathbf{x_i}$ and candidate sentences $\mathbf{s_j}$, we measure $\textbf{Score} (M, \mathbf{y_i}, \left[  \mathbf{s_j}; \mathbf{x_i}  \right]) = \mathrm{log} p_M(\mathbf{y}|\left[  \mathbf{s_j}; \mathbf{x_i}  \right])$, log likelihood assigned to target output according to LM $M$ when candidate sentence is prepended to the input. We consider the sentence with the highest log likelihood as a positive example $\mathbf{p_{i}}$ (line~\ref{alg:positive}). To construct negative examples $ \mathcal{N}_i = \{n_k\}_{k=1}^{5}$, we choose up to five sentences with top contriever score that has the log likelihood lower than the positive sentence for a threshold(line~\ref{alg:criteria}).

Training a compressor for QA task works similarly, but scoring will evaluate whether the LM will generate the correct answer with summary prepended (change in line \ref{alg:criteria}). Pseudo code for the QA tasks is in Figure~\ref{alg:contrastive_qa} the Appendix.
We train our encoder with a contrastive loss~\citep{karpukhin-etal-2020-dense}, maximizing the similarity between positive pairs $(\mathbf{x_i}, p_i)$ and minimize negative pairs $(\mathbf{x_i},  N_i)$. The training objective is to minimize $- log\frac{e^{sim(\mathbf{x_{i}}, p_{i})}}{e^{sim(\mathbf{x_{i}}, p_{i})} + \sum_{n_j \in N_i}e^{sim(\mathbf{x_{i}}, n_{j})}}$.

\paragraph{Data} For the language modeling task, we generate training data using the training split of the Wikitext-103 dataset, selecting the top 20 sentences from the top 5 BM25 retrieved documents for each input context $\mathbf{x}$. 
For the QA tasks, we generate training data using the training split and consider the top 20 sentences from the top 5 contriever-ms-marco \footnote{\url{https://huggingface.co/facebook/contriever-msmarco}} retrieved documents. We report detailed statistics for the training data in Table \ref{tab:training_data_stats} in the appendix.
For each sentence from the retrieved documents, we prepend the Wikipedia page title to it to for decontextualization.

\subsection{Abstractive Compression} \label{section:abstractive}
To train an abstractive compressor, we distill the query-focused summarization ability of extreme-scale LM by generating training dataset from it, filter the generated data, and train an encodedr decoder model from the filtered dataset~\citep{west-etal-2022-symbolic}. In contrast to prior work~\citep{jung2023impossible} which use intrinsic summarization metric for filtering, we use the LM's performance on the end task with the generated summaries prepended for filtering. Fig.~\ref{alg:cap} presents pseudo algorithm for training the abstractive compressor.

\subsubsection{Creating Training Dataset For Distillation}

\begin{wrapfigure}{r}{0.5\textwidth}
\vspace{-1.5em}
\begin{framed}
\small
\textbf{Input:} Teacher LM $M_{\text{t}}$, LM $M$, Summarization prompt set $\{\mathbf{p_i}\}_1^n$, Compressor $\mathtt{encdec}_\theta$, Training data $\{\mathbf{x_i, D_i, y_i}\}_1^T$ where $\mathbf{x_i}$ is input, $\mathbf{D_i}$ is the set of retrieved document for $\mathbf{x_i}$, $\mathbf{y_i}$ is the target answer. \\
\textbf{Output:} An updated $\mathtt{encdec}_\theta$\\
\begin{algorithmic}[1]
\STATE $\mathcal{T} \gets \emptyset$
\FOR{$i  \in \{1, \dots, T \}$} 
        \STATE{${v_r} \gets  -\infty $}
        \FOR{$j  \in \{1, \dots, n \}$} \label{alg:start_it}
         \STATE $\mathbf{s_j} = \textbf{Decode}(M_\text{t}, [\mathbf{p_j}; \mathbf{x_i};\mathbf{D_i}])$ \label{line:decode}
         \STATE{${v_j} = \textbf{Score}(M, \mathbf{y_i}, \left[  \mathbf{s_j}; \mathbf{x_i}  \right])$ }
         \IF{$ v_j > v_r $}
             \STATE{$\mathbf{s_t} \gets  \mathbf{s_j}$, ${v_r} \gets  v_j$}
         \ENDIF
        \ENDFOR \label{alg:end_it}
        \STATE{$v_d=\textbf{Score}(M, \mathbf{y_i}, \left[  \mathbf{x_i}  \right])$} 
         \IF{$v_r < v_d$} \label{alg:empty}
             \STATE{$T \gets T \cup \{(\mathbf{x_i, D_i,\emptyset})\} $}\label{alg:addempty}
             \STATE{break}
         \ENDIF
         \STATE{$T \gets T \cup \{(\mathbf{x_i,D_i,s_t})\} $}\label{alg:add}
    
\ENDFOR
\STATE $\mathtt{encdec}_\theta=\textbf{Finetune}(\mathtt{encdec}_\theta, T)$
\end{algorithmic}
\end{framed}\vspace{-0.8em}
\caption{Learning an abstractive compressor for language modeling task.}\label{alg:cap}
\vspace{-2.4em}
\end{wrapfigure}

\paragraph{Generation From Teacher Model} 

For the language modeling task, we manually construct four prompts to summarize evidence document set ($\{\mathbf{p_i}\}_1^n$).\footnote{The exact prompts can be found in \autoref{tab:compression_example} in \ref{sec:training_detail}.} Given an input $\mathbf{x_i}$, a retrieved document set $\mathbf{D}_i$, and a prompt $\mathbf{p_j}$ to summarize the document set with respect to the input, GPT-3.5 \footnote{We use \texttt{gpt-3.5-turbo} in all our experiments.} generates a summary (line~\ref{line:decode}).

\paragraph{Filtering with Critic} After generating a summary for each prompt template, we select the summary which results in the highest end task performance for each example ($\mathbf{s_t}$) as the target summary (line~\ref{alg:start_it}-\ref{alg:end_it}). $\textbf{Score} (M, \mathbf{y_i}, \left[  \mathbf{s_j}; \mathbf{x_i}  \right])$ is the same as the extractive compressor above. We then compare the end task performance with the target summary prepended and with input $\mathbf{x_i}$ only (i.e. no retrieval) on base model $M$ (line~\ref{alg:empty}). If the end task performance gets worse (e.g., increase in perplexity) when prepending the summary, we set the target summary to an empty string (line~\ref{alg:addempty}), otherwise we add the target summary to the training set (line~\ref{alg:add}). This allows for selective augmentation and mitigates the risk of prepending irrelevant documents.

Constructing training datasets for the question answering tasks works similarly, with the following modifications. As summarization for the question answering task is more straightforward, we use a single prompt for each dataset. We filter out examples where prepending the summary does not lead to performance improvement. Pseudo code for the QA tasks is in Figure~\ref{alg:cap_qa} in the Appendix.

\paragraph{Model \& Training} We use encoder-decoder LM (775M), initialized from T5-large checkpoint~\citep{raffel2020exploring}. This model has been trained with summarization datasets~\citep{Hermann2015TeachingMT}.

\paragraph{Data} We summarize top 5 retrieved documents for both language modeling and question answering tasks. We generate training examples using 2\% of the training set for the Wikitext-103 dataset. We generate training examples from the entire NQ training set and TriviaQA training set. For HotpotQA, we only generate summaries for the training data where the gold answer is in the retrieved documents (56\% of the training data) to reduce API costs.  We report percentage of data filtered and percentage of empty summaries in Table \ref{tab:training_data_stats} in \ref{sec:training_data}.

\section{Experimental Settings}

We evaluate our approach on language modeling and open-domain QA following prior work \citep{Shi2023REPLUGRB, Ram2023InContextRL}. For both tasks, we report the task performance as a measure of effectiveness and the number of tokens provided in context as a measure of efficiency.

\subsection{Language Modeling}\label{sec:lm_exp}

We evaluate language modeling perplexity on {WikiText-103} \citep{Merity2016PointerSM} benchmark on three open-sourced LMs of varying scale: GPT2 (117M), GPT2-XL (1.5B; \citet{Radford2019LanguageMA}) and GPT-J (6B; \citet{gpt-j}). We train our compressors using GPT2 as the base model and evaluate whether the trained compressor transfer to GPT2-XL and GPT-J. We use the BM25 retriever~\citep{Robertson2009ThePR} to retrieve from the Wikipedia corpus from Dec. 20, 2018 \citep{karpukhin-etal-2020-dense}. The articles are then truncated into non-overlapping documents of 100 words. During retrieval, articles containing the input sequence $\mathbf{x}$ is removed from the corpus to prevent data contamination. Following~\cite{Ram2023InContextRL}, we perform retrieval every 32 tokens.

\subsection{Open-domain QA}\label{sec:nq_exp}
\paragraph{Datasets} We evaluate our model on three benchmark dataset: Natural Questions (NQ) \citep{kwiatkowski2019natural}, TriviaQA \citep{triviaqa}) and HotpotQA \citep{yang2018hotpotqa}. We report results on development set of NQ, test set of TriviaQA and randomly sampled 500 examples from HotpotQA development set. We report Exact Match (EM) and F1token-level F1 of answer strings to measure end task performance.

\paragraph{Base Language Models \& Retrieval Corpus} We use Flan-UL2 (20B)\citep{Chung2022ScalingIL}, a large scale instruction-tuned LM. We use contriever model trained on MS MARCO dataset~\citep{Campos2016MSMA} as a retriever on Wikipedia corpus from Dec. 20, 2018 for all three datasets. The articles are truncated into non-overlapping documents of 100 words.

\paragraph{Prompt Format} We include few-shot in-context examples in the prompt, followed by the retrieved documents and the question. We use five randomly sampled training examples as in-context examples, which constitutes 110, 147, and 149 tokens on average for NQ, TQA and HotpotQA respectively. For retrieved documents, we concatenate them in ascending order of retrieval score, with the highest scored document closest to the question \citep{Si2022PromptingGT}. We do not include the retrieved documents for in-context examples as it did not improve performance. An example input can be found in Appendix Table \ref{tab:icl_prompt}.

\subsection{Baselines and Oracles}

 \paragraph{Baselines} We first consider two heuristic token and phrase-level compression methods: \textbf{BoW}, which converts the retrieved documents to a list of ordered unigram and concatenates them together and \textbf{Named Entities (NE)}, which extracts a list of ordered named entities from retrieved documents and concatenates them. For the extractive compressor on the language modeling task, we use \textbf{BM25} and \textbf{Contriever} \cite{izacard2021contriever}, which rank the sentences by their similarity to the input $\mathbf{x}$ as baselines. For the QA datasets, we report results using \textbf{BM25}, \textbf{Contriever} finetuned on MS MARCO and \textbf{DPR} \citep{Karpukhin2020DensePR} fine-tuned on NQ. We also report a \textbf{Random} baseline which randomly selects a sentence from the retrieved documents. For abstractive compression, we report the performance of the off-the-shelf \textbf{T5 (large, 770M)} model and that of \textbf{GPT-3.5} model. As we experimented with multiple prompts for the language modeling task, we report the performance of the summaries generated by \textbf{GPT-3.5} model with the best single prompt.

\paragraph{Oracle} We explore the performance upper bound of compressioin by considering two oracle approaches. For the extractive approach, we construct oracle compressor by considering all sentences $s_i$ in the evidence document set and choosing the sentence that leads to the best end task performance (i.e., lowest perplexity or highest answer accuracy) for each example. For the abstractive approach, we consider summaries generated from different prompts ($\{{\mathbf{s_j}}\}_1^n$ in Figure~\ref{alg:cap}) and empty summary, and choose the one that leads to the best end task performance. As oracle compression is model dependent, we also report model-independent results by always using GPT-2 as a reference LM (\textit{Oracle w/ gpt2}) to test how well oracle sentences for one model transfer to other models for the language modeling task.

\section{Results}

\begin{table*}
    \vspace*{-6mm}
\caption{Results on language modeling task. We report results on GPT-2, GPT2-XL and GPT-J with compressors trained with GPT-2.}
\footnotesize
\setlength{\tabcolsep}{4.5pt}
\begin{center}
\begin{tabular}{@{}lcccccc@{}}
\toprule

 & \multicolumn{2}{c}{\textbf{In-Domain}} & \multicolumn{4}{c}{\textbf{Out-Domain}} \\
 & \multicolumn{2}{c}{\textbf{GPT2 (117M)}} & \multicolumn{2}{c}{\textbf{\textit{GPT2-XL (1.5B})}} & \multicolumn{2}{c}{\textbf{\textit{GPT-J (6B)}}} \\
\textbf{In-context Evidence} & \textbf{\# tokens} & \textbf{PPL} & \textbf{\# tokens} & \textbf{PPL} & \textbf{\# tokens} & \textbf{PPL} \\
\midrule
- & 0 & 37.84 & 0 & 19.89 & 0 & 11.44 \\ 
\midrule
\multicolumn{7}{l}{\textit{\textbf{RALM without compression}}} \\
Top 1 document & 141 & 32.90 & 14 & 17.86 & 141 & 10.57 \\ 
Top 5 documents & 512 & 35.53 & - & - & - & - \\ 
\midrule
\multicolumn{7}{l}{\textit{\textbf{Phrase/token level compression}}} \\
Top 1 document (BoW) & 66 & 36.13 & 66 & 18.85 & 66 & 10.97 \\ 
Top 1 document (NE) & 34 & 37.23 & 33 & 19.67 & 33 & 11.39 \\ 
\midrule
\multicolumn{7}{l}{\textit{\textbf{Extractive compression of Top 5 documents (select top 1 sentence)}}} \\
\rowcolor{LightGrey}
\textit{Oracle}  & 32 & 30.36 & 32 & 16.58 & 31 & 9.92  \\ 
\rowcolor{LightGrey}
\textit{Oracle (w/ gpt2)}  & 32 & 30.36 &  32 & 16.99 & 32 & 10.22 \\ 
                       Random & 27 & 36.98 & 27 & 19.55 & 27 & 11.32 \\  
                       BM25 & 33 & 36.63 & 33 & 19.02 & 33 & 11.08 \\ 
                       Contriever & 33 & 35.54 & 33 & 18.98 & 33 & 11.05 \\ 
                      
                       Ours (init. w/ Contriever) & 31 & {33.67} & 31 & {18.19} & 31 & {10.73} \\
\midrule
\multicolumn{7}{l}{\textit{\textbf{Abstractive compression of Top 5 documents}} } \\
 
\rowcolor{LightGrey}\textit{Oracle}  &  68 & 30.67 & 66 & 16.87 &  65 & 10.10 \\
 
\rowcolor{LightGrey} \textit{Oracle (w/ gpt2)} & 68 & 30.67 & 68 & 17.23 & 68 & 10.37  \\ 

\textit{GPT-3.5} & 33 & 34.84 & 33 & 18.70 & 33 & 10.96 \\ 

T5 & \textbf{15} & 37.80  & \textbf{15}  & 19.92 & \textbf{15} & 11.5 \\

Ours (init. w/ T5) & 15 & \textbf{33.64} & 15 & \textbf{18.09} & 15 & \textbf{10.66}\\

\bottomrule
\end{tabular} 
\end{center}\vspace{-0.6em}
\label{tab:lm_results}
\vspace{-0.1in}
\end{table*}

\paragraph{Language modeling} Table \ref{tab:lm_results} reports the results on language modeling task. \textit{All} retrieval augmentation methods improve perplexity over no retrieval setting across three LMs. Heuristic token / phrase-level compression methods (BoW and NE) are worse than prepending uncompressed documents, potentially due to the disfluency of the prepended text.

Both oracle settings show substantial gain over prepending the entire document set, with only 6-13\% of tokens. More tokens are not always better: prepending top 1 document outperforms prepending top 5 documents. This confirms that the naive retrieve-and-prepend approach has a significant room for improvement, as prepending irrelevant documents can hurt performances.

Our trained extractive compressor significantly outperforms other extractive baselines (Contriever and BM25) across all three LMs, while prepending slightly fewer tokens. Comparing to prepending one document, we achieve a compression ratio of 25\% at minimum performance drop. Our trained abstractive compressor performs the best across the board, achieving the lowest perplexity \textbf{and} the highest compression ratio. Our abstractive compressor achieves high compression rate through selective augmentation, prepending summaries to only 33\% of examples (length distribution of generated summaries in Fig. \ref{fig:length_hist}).

\begin{table}
\vspace{-1em}
\caption{Open-domain QA results with Flan-UL2 (20B) as  the LM $M$. We report number of tokens provided as in-context evidence document, excluding the in-context examples. We train separate compressors (one extractive, one abstractive) for each dataset. Extractive compressor selects one sentence for NQ/TQA, and two sentences for HotpotQA.}
\vspace{-0.5em}
\small
\begin{center}
\begin{tabular}{@{}lcccccccccc@{}}
\toprule
 &  \multicolumn{3}{c}{\textbf{NQ}} & \multicolumn{3}{c}{\textbf{TQA}} & \multicolumn{3}{c}{\textbf{HotpotQA}}\\
\textbf{In-Context evidence} & \textbf{ \# tok} & \textbf{EM} & \textbf{F1} & \textbf{\# tok} & \textbf{EM} & \textbf{F1} & \textbf{\# tok} & \textbf{EM} & \textbf{F1} \\
\midrule
        - & 0  & 21.99 & 29.38 & 0 & 49.33 & 54.85 & 0 & 17.80 & 26.10 \\ 

\midrule
\multicolumn{10}{l}{\textit{\textbf{RALM without compression}}} \\
       Top 1 documents & 132  & 33.07 & 41.45 & 136 & 57.84 & 64.94 &  138 & 28.80 & 40.58 \\
           {Top 5 documents} & 660 & \textbf{39.39} & \textbf{48.28} & 677 & \textbf{62.37} & \textbf{70.09} & 684  & \textbf{32.80} & \textbf{43.90} \\
\midrule
\multicolumn{10}{l}{\textit{\textbf{Phrase/token level compression}}} \\
    Top 5 documents (NE) &  338 & 23.60 & 31.02 & 128 & 54.96 & 61.19 &  157 & 22.20 & 31.89\\
      {Top 5 documents (BoW) } & 450 & 28.48 & 36.84 & 259 & 58.16 & 65.15 & 255 & 25.60 & 36.00 \\
\midrule
\multicolumn{10}{l}{\textit{\textbf{Extractive compression of top 5 documents}}} \\
  \rowcolor{LightGrey}           \textit{Oracle} & 34 & 60.22 & 64.25 &  32 & 79.29 & 82.06 & 70 & 41.80 & 51.07 \\ 
     Random & 32 & 23.27 & 31.09 &  31 & 50.18 & 56.24 &  61 & 21.00 & 29.86 \\ 
                BM25 & 36 & 25.82 & 33.63 & 37 & 54.67 & 61.19 & 74 & 26.80 & 38.02 \\ 
                
         DPR & 39 & 34.32 & 43.38 & 41 & 56.58 & 62.96 & 78 & 27.40 & 38.15 \\ 
        Contriever & 36 & 30.06 & 31.92 & 40 & 53.67 & 60.01 & 78  & 28.60 & 39.48 \\
         Ours & 37 & {36.57} & {44.22} & 38 & \textbf{58.99} & {65.26} & 75 &\textbf{30.40} & \textbf{40.14} \\
\midrule 
\multicolumn{10}{l}{\textit{\textbf{Abstractive compression of top 5 documents}}} \\
     \rowcolor{LightGrey}    \textit{Oracle}  & 51 & 45.68 & 53.66 &  37 & 71.01 & 76.38 & 102 & 35.80 & 46.25 \\ 
         \textit{GPT-3.5} & 56 & 37.12 & 46.35 & 41 & 62.03 & 69.66 & 107 & 31.60 & 42.65 \\ 
        T5 & \textbf{10} & 25.90 & 34.63 & \textbf{7} & 55.18 & 62.34 & \textbf{7} & 23.20 & 33.19 \\ 
        Ours & 36 & \textbf{37.04} & \textbf{45.47} & 32  & 58.68 &  \textbf{66.34} & 64 & {28.20} & {37.91} &  \\ 
        
\bottomrule
\end{tabular} 
\end{center}\vspace{-0.8em}
\label{tab:nq_results}
\end{table}

\paragraph{Open-domain QA} We report the results on QA tasks in Table \ref{tab:nq_results}. Similar to the language modeling task, all retrieval augmentation methods improve performance over no retrieval setting, across three datasets, consistent with previous study on other LMs \citep{Shi2023REPLUGRB, Mallen2022WhenNT, Si2022PromptingGT}. Unlike language modeling, prepending five documents shows significant gains over prepending a single document, motivating the use of compression to incorporate more documents.

We find that extractive oracle outperforms the abstractive one in all datasets. Extractive oracle selects the best one from $N$ candidate sentences, while abstractive oracle selects from two options -- prepending GPT-3.5 summary or prepending nothing. Both oracles show improvements over prepending all information, suggesting that removing irrelevant information benefit the model.\footnote{We provide an example where our compressed summary yields correct answer while prepending full document does not in Table \ref{tab:case_study} in the appendix.}

Among extractive baselines, DPR performs the best as it has been fine-tuned on high-quality NQ data. On NQ, selecting the top 1 DPR ranked sentences from top 5 documents outperforms prepending top 1 document, with much fewer tokens (39 vs. 132). However, its performance degrades in out of domain datasets. Off-the-shelf summarization model (T5) boasts the highest level of compression, achieving 4-6 points gains in EM while adding mere 7-10 tokens.

The trained compressors, both extractive and abstractive, shows promising performances. On NQ and TQA, the abstractive approach is more effective. On NQ, it achieves a compression ratio of 5\% tokens while losing 2 EM points compared to prepending full documents. On TQA, we observe similar trends, compression ratio of 5\% tokens while losing 3.7 EM points compared to prepending full sets of documents. 
On HotpotQA that requires multihop understanding of documents, we find extractive approach to be more helpful, achieving 11\% compression rate while losing 2.4 EM points compared to prepending full documents. We find that learning an abstractive compressor for more complex tasks, such as HotpotQA, demands further study. While extreme-scale LLM boasts competitive summarization performance under single document setting, they are not good at synthesizing information from multiple documents (\citet{shaib-etal-2023-summarizing} and hallucinate more often; See Section~\ref{sec:analysis} for further analysis).

\section{Analysis and Discussions}\label{sec:analysis}

\paragraph{Transferring Across Different LMs}
One benefit of textual summary is that they can transfer to other LMs, unlike approaches such as soft prompts ~\citep{wingate-etal-2022-prompt, chevalier2023adapting, Mu2023LearningTC}. We evaluate whether our compressors trained to achieve high performance with respect to a specific LM (GPT2 for language modeling, FlanUL2 for open domain QA) can transfer to other LMs. For language modeling, we find that trained compressor transfers well to other LMs (GPT2-XL and GPT-J), despite they are much larger LMs (Table \ref{tab:lm_results}. For open domain QA, we tested transferring our compressors to LLaMA-13B \citep{Touvron2023Llama2O} model. The results can be found in Table~\ref{tab:llama2_results} in the appendix. Overall, the performance is worse than the LM from which compressors are trained on, sometimes unable to outperform other compression baselines (e.g., no clear gain from using contriever vs. our trained contriever on TQA/HotpotQA), leaving considerable gap to the oracle compressions for LLAMA itself. Yet, on NQ/TQA, our compressor obtains 5\% compression ratio with less than 5 EM drop compared to full document setting, showing the robustness of our retrieve-compress-prepend paradigm.

\begin{wrapfigure}{l}{0.4\textwidth}
\vspace{-1.5em}
    \centering
    \includegraphics[width=0.4\textwidth]{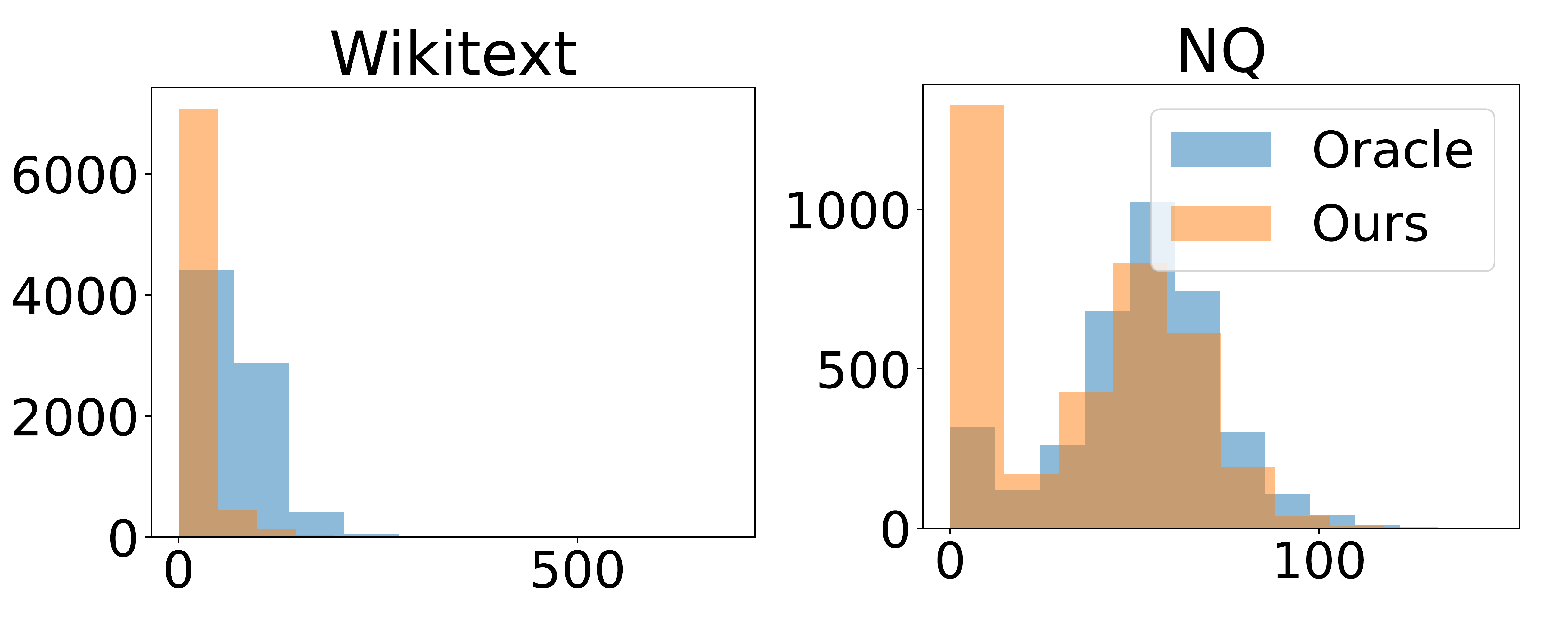}
    \vspace{-0.25in}
    \caption{Histogram of abstractive summary length (\# tokens) distribution.}
    \label{fig:length_hist_main}
\vspace{-0.4em}
\end{wrapfigure}
\paragraph{How do the length of the summaries vary?} 
Can the learned compressor reliably determine when LMs require retrieved documents or not? 
As retrieved documents were \textbf{hurting} the model performances for some input queries, 4-24\% of training examples for abstractive compressors contain empty summary. Fig.~\ref{fig:length_hist_main} presents the length distribution of abstractive summaries on NQ and Wikitext (histograms for other datasets is in Fig.~\ref{fig:length_hist} in the appendix). The input document lengths do not vary significantly across examples, yet we find the abstractive summary vary significantly in length, suggesting abstractive compressor enables selective retrieval augmentation.
We have not experimented selective compression with extractive compressor, fixing the number of prepended sentences for the entire dataset (1 for Wikitext, 1 for NQ/TQA, 2 for HotpotQA). Allowing adaptive augmentation with extractive summarizer can be a promising direction for future work. 

\paragraph{How does model leverage the in-context documents?}

\begin{wraptable}{r}{0.5\textwidth}\vspace{-1em}
\caption{Analysis on in-context evidence to answer questions in NQ dev set. For the last column, we report how frequently model copies from its evidence on (1) a subset where gold answer is in the evidence document / (2) when it is not. }\vspace{-0.5em}
\small
\begin{center}
\begin{tabular}{@{}lccc@{}}
\toprule
  {\textbf{Evidence}} &{\textbf{EM}}    &{\textbf{\%Gold in Evi.}} & {\textbf{\%Pred in Evi.}}\\\midrule
  Top 1 &  33.1  & 36 & 92 / 51 \\
  Top 5 & 39.3 & \textbf{57} & 96 / 81 \\ \midrule
  NE  & 26.0  & {46} & 84 / 48 \\ 
\rowcolor{LightGrey}  Oracle sent & 60.2& 34 & 93 /  16 \\
  Contriever & 30.2 & 25 & 88 / 36 \\
  Ours  &  36.6& 28 & 90 / 33 \\ \midrule
  GPT-3.5 & 37.1  & 45 & \textbf{98} / \textbf{85} \\ 
  T5 & 25.9  & 30 & 52 / 20 \\ 
  Ours & 37.0  & 34 & \textbf{98} / 39 \\
\bottomrule
\end{tabular} 
\end{center}\vspace{-0.5em}
\label{tab:incontext_analysis}\vspace{-0.6em}
\end{wraptable}

We evaluate whether retrieval augmented LMs tend to copy answers verbatim from in-context evidence documents or generate answers not present in the documents. This is an desired behavior \textit{only} when the gold answer is in the evidence. We first report how frequently a gold answer span is present in evidence text (\% Gold in Evi). As expected, full documents contain the answer most frequently, followed by NE and GPT-3. However, having more gold answers in the evidence doesn't equate to better performance, as the model cannot always identify the correct answer from the evidence (84 \% for \textbf{NE}  v.s. 98\% for \textbf{T5(ours)}).

We also observe that model can be easily distracted by irrelevant contexts, copying from a document span even when it does not contain gold answer, echoing findings from prior work \citep{Shi2023LargeLM}. Prepending top 5 documents has a higher frequency (81\%) of copying incorrectly compared to top 1 document (51\%), and GPT-3 compression leads to an even higher incorrect copying frequency (85\%), potentially as query-focused summarization generates sentences that seemingly contains the answer. Our compressor successfully reduce such erroneous behavior to 39\%.

\paragraph{Is generated summary faithful and comprehensive?}

We (the authors) manually evaluate outputs of the abstractive compressors on two axes~ \citep{Chen2023ComplexCV}: \textbf{Faithfulness}: whether the summary can be entailed by the retrieved documents, \textbf{Comprehensiveness}:  whether the summary contains sufficient information to answer the question, regardless of whether the generated information comes from the retrieved documents. For both, we select one of three labels: \textbf{Y}es, \textbf{P}artially, \textbf{No}, {and report the \% of \textbf{Useful} summaries which are both faithful and comprehensive}. Annotation sample can be found in Table \ref{tab:example_summaries} in the appendix. 
We evaluate the summaries generated by GPT-3.5 and our abstractive compressor. We randomly sample 30 non-empty summaries from the test set.

\begin{wraptable}{r}{0.60\textwidth}
\vspace{-1em}
\caption{Manual analysis on abstractive summaries generated for NQ, TQA and HotpotQA (HQA) dataset. }
\begin{center}
\small
\begin{tabular}{@{}llccccccc@{}}
\toprule
  \multirow{2}{*}{\textbf{Dataset}} & \multirow{2}{*}{\textbf{Model}} &  \multicolumn{3}{c}{\textbf{\% Faithful}} & \multicolumn{3}{c}{\textbf{\%  Compre.}} & \multirow{2}{*}{\textbf{\% Use.}}\\
  & & Y & P & N & Y & P & N \\ 
  \midrule
  \multirow{2}{*}{NQ} & GPT-3.5& 90 & 0 & 10 & 97 & 0 & 3 & 83 \\ 
                     & Ours &  80 & 13 & 7 &  100 & 0 & 0 & 80 \\ 
  \multirow{2}{*}{TQA} & GPT-3.5 & 97 & 0 & 3 & 90 & 0 & 10 & 83 \\ 
                               & Ours & 83 & 3 & 14 & 96 & 0 & 4 & 77 \\ 
  \multirow{2}{*}{HQA} & GPT-3.5 & 74 & 0 & 26 & 78 & 0 & 22 & 50 \\
                               & Ours & 67 & 0 & 33 & 74 & 0 & 26 & 40\\ 
\bottomrule
\end{tabular} 
\end{center}\vspace{-0.3em}
\label{tab:manual_analysis}\vspace{-0.3em}
\end{wraptable}

Table \ref{tab:manual_analysis} presents annotation results. GPT-3.5, substantially bigger than our compressor, generates more useful summary across all three datasets. Overall, our abstractive compressors were less faithful compared to the original GPT-3.5, while improving comprehensiveness. The effectiveness of summarization also depends on the datasets -- summaries from both models were the most faithful for TQA and the least faithful for HotpotQA dataset. In terms of comprehensiveness, we find both models easily find the information for NQ, but struggle with HotpotQA. These results partially explain why the performance gain was limited for HotpotQA.

\section{Related Work}

\paragraph{Efficient RALM} ~\citet{he-etal-2021-efficient} improves efficiency of RALMs by improving retrieval components, such as data store compression, dimensionality reduction for neural retriever. A line of work also introduces reducing retrieval frequency through selective retrieval \citep{he-etal-2021-efficient,Mallen2022WhenNT} or using a larger stride \citep{martins-etal-2022-chunk}. In this work, we improve efficiency of RALM by compressing retrieved documents into a concise summary or an empty sequence, facilitating  selective retrieval augmentation.

\paragraph{Prompt Compression}
Recent work~\citep{wingate-etal-2022-prompt, chevalier2023adapting, Mu2023LearningTC} proposes compressing long contexts into summary vectors (soft prompts) that can be used by LMs, rather than shorter textual summaries. Such soft prompts can serve as efficient replacements for plain-text demonstrations, minimizing the computational costs during inference. Another related line of work proposes context distillation \citep{Snell2022LearningBD,Choi2022PromptIP,Padmanabhan2023PropagatingKU}, which injects the prepended context into the parameters of an LM. Compared to above approaches, our approach yields more interpretable textual summary that can transfer across different LMs, and can be applied to black box LMs without requiring gradient updates. Prior work has studied textual compression for other tasks, such as political fact checking ~\citep{Chen2023ComplexCV} and instruction learning \citep{Yin2023DidYR}.

\paragraph{Distillation / Goal Oriented Summarization}
 Recent work introduces symbolic knowledge distillation~\citep{west-etal-2022-symbolic}, which transfers knowledge from a teacher model by generating a training dataset with the teacher model and train a student model on it. For better performance, they introduce critic criteria, which filter undesirable examples from generated training dataset. Such distillation technique has been applied for various applications including summarization~\citep{jung2023impossible}, which aims to generate high quality summaries while we optimize for generating effective summary for downstream LMs. One work that is similar to our setting is \citet{hsu-tan-2021-decision} which trains an extractive summarization model to optimize for prediction accuracy of a sentiment prediction model based on the summary.

\section{Conclusion}
We introduce \modelnospace, a method which compresses retrieved documents into textual summaries before prepending them to improve in-context retrieval augmented language models. We present two compression models -- an extractive compressor and an abstractive compressor. We design a training scheme which leverages end task signals from a blackbox LM to generate useful summaries and allowing the compression models to perform selective augmentation. Our experiments show that our compressors can improve the efficiency of retrieval augmented LMs significantly with minimal drop in performances.

\section*{Acknowledgement}
We thank the members of the UT and UW NLP community for feedback on the project. We especially thank Alisa Liu, Junyi Jessy Li and Greg Durrett for providing comments on the draft. The project is partially funded by NSF grant (IIS-2312948).

\section*{Ethics Statement}
We use commercial language model to generate training data for our compressors, which might include factual error. We conduct careful human evaluation on the data generated and present our analysis in the paper.

\section*{Reproducibility Statement}
We release our codes, prompt, and data generated with API access publicly.

\bibliography{iclr2024_conference}
\bibliographystyle{iclr2024_conference}
\newpage
\appendix
\section{Appendix}

\subsection{Compressor training data generation}\label{sec:training_data}

We report the statistics of the data used to train compressors in Table \ref{tab:training_data_stats}. We use SpaCy \citep{spacy2} to extract named entities.

\paragraph{Extractive Data Generation} We generate data using the training data for the four datasets we tested (Wikitext, NQ, TQA and HotpotQA).  We use the NLTK package to perform sentence splitting. We remove examples without any negatives.

\paragraph{Abstractive Data Generation} We report prompt used to generate summaries in Table \ref{tab:wikitext_prompt}. We queried the Open AI API with temperature of 0.7 and {top p} = 1. For the language modeling task, we use an ensemble of four prompts and choose the one which leads to the lowest perplexity as the target. If none of the summaries lead to perplexity decrease, we treat an empty summary as target. We queried the OpenAI API with temperature of 0.7 and {top p} = 1. We generate four summaries per example for randomly sampled 2\% of the training data (48,013 examples).

\begin{figure}
    \centering
    \includegraphics[width=\textwidth]{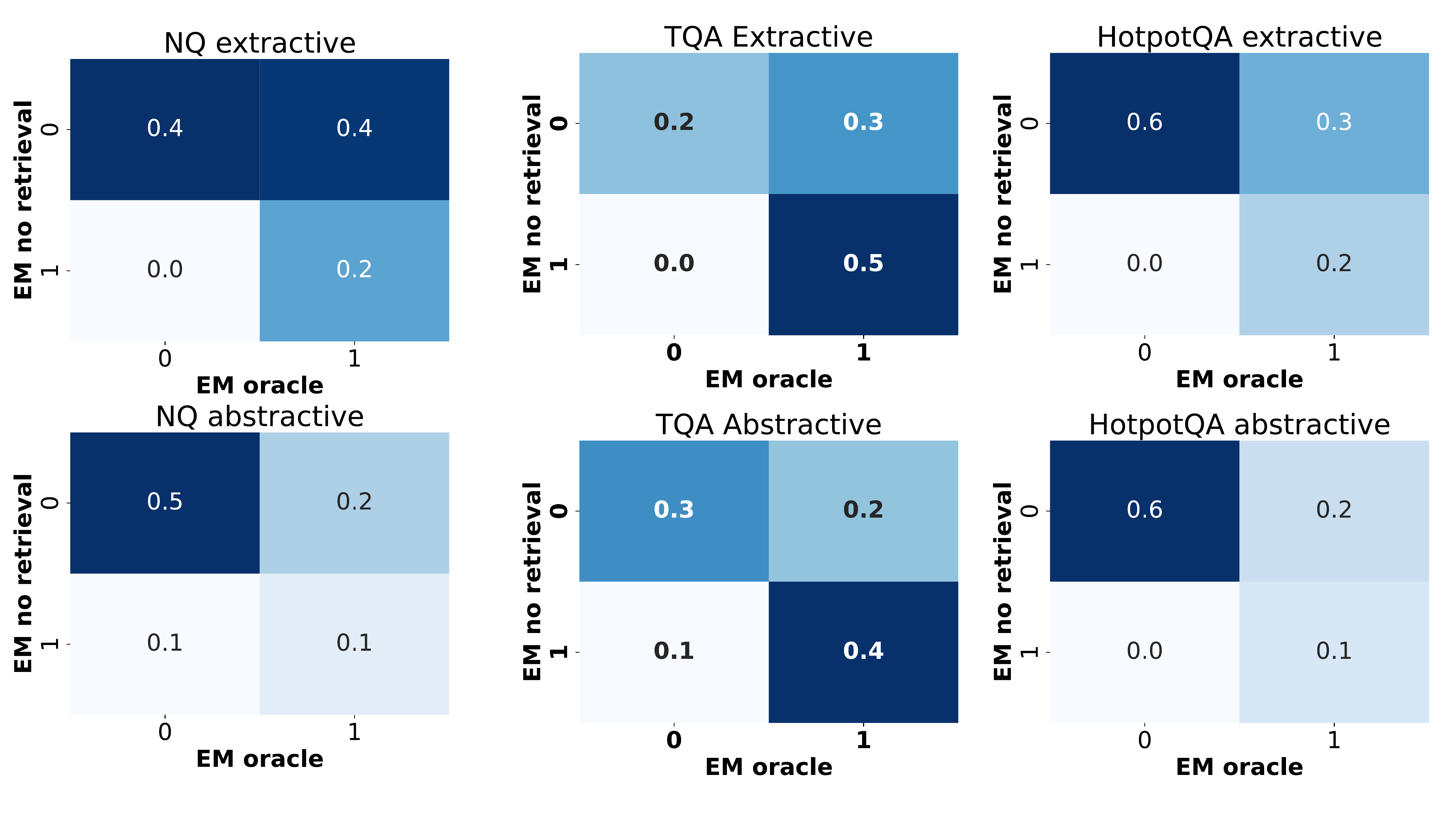}
    \caption{We report the data distribution on NQ dev set, TriviaQA dev set and HotpotQA dev set comparing the end task performance when prepending the oracle compression method (oracle sentence or GPT-3 summaries) and when not prepending anything for the base model (Flan-UL2).}
    \label{fig:oracle_analysis_heatmap}
    
\end{figure}

\begin{figure}
\begin{framed}
\small
\textbf{Input:} Base LM $M$, Compressor encoder $\mathtt{enc}_\theta$, 
Training data $\{\mathbf{x_i, S_i, y_i}\}_1^T$ where $\mathbf{x_i}$ is input, $\mathbf{S_i}=\{\mathbf{s_j}\}_1^n$ is a set of candidate sentences from the retrieved document for $\mathbf{x_i}$, $\mathbf{y_i}$ is the target answer.

\textbf{Output:} An updated extractive compressor encoder $\texttt{enc}_\theta$


\begin{algorithmic}[1]
\STATE $\mathcal{T} \gets \emptyset$

\FOR{$i  \in \{1, \dots, T \}$} 
    \STATE $\mathbf{p_{i}} \gets \mathrm{argMax}_{\mathbf{s_j} \in \{\mathbf{S_i}\}} \textbf{Score}(M, \mathbf{y_i}, \left[  \mathbf{s_j}; \mathbf{x_i}  \right])$ \label{alg:positive}

    \FOR{$j  \in \{1, \dots, n \}$} 
     \STATE $\mathcal{L} \gets \emptyset $
        \IF{$\textbf{Score}(M, \mathbf{y_i}, \left[  \mathbf{s_j}; \mathbf{x_i}  \right]) < \textbf{Score}(M, \mathbf{y_i}, \left[  \mathbf{p_i}; \mathbf{x_i}  \right]) $}\label{alg:criteria}
            \STATE{$\mathcal{L} \gets \mathcal{L} \cup \mathbf{s_i}  $ }    
    \ENDIF
    \ENDFOR
    \IF{$|\mathcal{L}|>0$}
    \STATE $\mathcal{N}_i \gets \mathrm{argTop}5_{\mathbf{s_j} \in \mathcal{L}} (\langle \mathtt{enc}_\theta(\mathbf{s_j}), \mathtt{enc}_\theta(\mathbf{x_i}) \rangle) $

\STATE $\mathcal{T} \gets \mathcal{T} \cup \{ (\mathbf{x_i, p_i}, \mathcal{N}_i) \}$
\ENDIF
\ENDFOR



\STATE $\mathtt{enc}_\theta=\textbf{Finetune}(\mathtt{enc}_\theta, \mathcal{T})$

\end{algorithmic}
\end{framed}
\vspace{-0.5em}
\caption{Learning an extractive compressor for QA task. The \textbf{Score} here is the exact match between the decoded answer and the gold answers.}\label{alg:contrastive_qa}
\end{figure}
\begin{figure}
\begin{framed}
\small
\textbf{Input:} Teacher LM $M_{\text{t}}$, Base LM $M$, Summarization prompt $p$, Compressor $\mathtt{encdec}_\theta$, Training data $\{\mathbf{x_i, D_i, y_i}\}_1^T$ where $\mathbf{x_i}$ is input, $\mathbf{D_i}$ is the set of retrieved document for $\mathbf{x_i}$, $\mathbf{y_i}$ is the target answer. \\
\textbf{Output:} An updated $\mathtt{encdec}_\theta$\\
\begin{algorithmic}[1]
\STATE $\mathcal{T} \gets \emptyset$
\FOR{$i  \in \{1, \dots, T \}$} 
         \STATE $\mathbf{s_i} = \textbf{Decode}(M_\text{t}, [\mathbf{p}; \mathbf{x_i};\mathbf{D_i}])$ \label{line:decode}
        \STATE{${v_s} = \textbf{Score}(M, \mathbf{y_i}, \left[  \mathbf{s_i}; \mathbf{x_i}  \right])$ }
         
        \STATE{$v_d=\textbf{Score}(M, \mathbf{y_i}, \left[  \mathbf{x_i}  \right])$} 
         \IF{$v_s < v_d$} \label{alg:empty}
             \STATE{$T \gets T \cup \{(\mathbf{x_i, D_i,\emptyset})\} $}\label{alg:addempty}
             \STATE{break}
         \ENDIF
         \STATE{$T \gets T \cup \{(\mathbf{x_i,D_i,s_i})\} $}\label{alg:add}
    
\ENDFOR
\STATE $\mathtt{encdec}_\theta=\textbf{Finetune}(\mathtt{encdec}_\theta, T)$
\end{algorithmic}
\end{framed}
\caption{Learning an abstractive compressor for QA task. The \textbf{Score} here is the exact match between the decoded answer and the gold answers.}\label{alg:cap_qa}
\end{figure}

\subsection{Compressor Training Details}\label{sec:training_detail}

\paragraph{Extractive Compressor} 
For language modeling, we use the contriever checkpoint \footnote{\url{https://huggingface.co/facebook/contriever}} trained with unsupervised data. For the QA tasks, we use the contriever checkpoint fine-tuned on the MSMARCO task~\citep{Campos2016MSMA} \footnote{\url{https://huggingface.co/facebook/contriever-msmarco}} , following prior work \citep{Si2022PromptingGT, Shi2023REPLUGRB}.
We implement the model using the Transformers \citep{Wolf2019HuggingFacesTS} and the sentence-transformer library \citep{reimers-gurevych-2019-sentence}. We train with Adam optimizer \citep{Kingma2014AdamAM}, using a batch size of 64, learning rate of 2e-5 and 1000 warmup steps for 3 epochs. We report results on the model with the best reranked perplexity on our validation set for the language modeling task and the best reranked accuracy for the QA tasks.

\paragraph{Abstractive Compressor}
We implement the model using the Transformers \citep{Wolf2019HuggingFacesTS}. We train abstractive summarizer with Adam optimizer \citep{Kingma2014AdamAM}, using a batch size of 16, learning rate of 1e-5 and 1000 warmup steps for 3 epochs. 

\begin{table}
\caption{Training data statistics for abstractive and extractive compressors.}
\small
\begin{center}
\begin{tabular}{@{}lccccccccc@{}}
\toprule
  \multirow{2}{*}{\textbf{Dataset}} &  \multicolumn{4}{c}{\textbf{Extractive}} & \multicolumn{4}{c}{\textbf{Abstractive}}\\
  & Train & Validation & \% filtered & $|\mathcal{N}|$ & Train & Validation & \% filtered & \% empty \\ 
  \midrule
  NQ & 42,149 & 9,769 & 46 & 4.44 &  39,466 & 4,931 & 50 & 25 \\ 
  TQA & 70,032 & 8,753 & 56 & 4.37& 48,322 & 5,887 & 32 & 16 \\ 
  HotpotQA & 24,526 & 3,068 & 69 & 4.33 &  26,556 & 2,937 & 42 & 4 \\
  Wikitext & 1,398,318 & 1,5483 & 41 & 4.04 & 38,410 & 9,603 & 0 & 24\\ 
\bottomrule
\end{tabular} 
\end{center}\vspace{-0.3em}
\label{tab:training_data_stats}
\end{table}

\begin{figure}
    \centering
    \includegraphics[width=\textwidth]{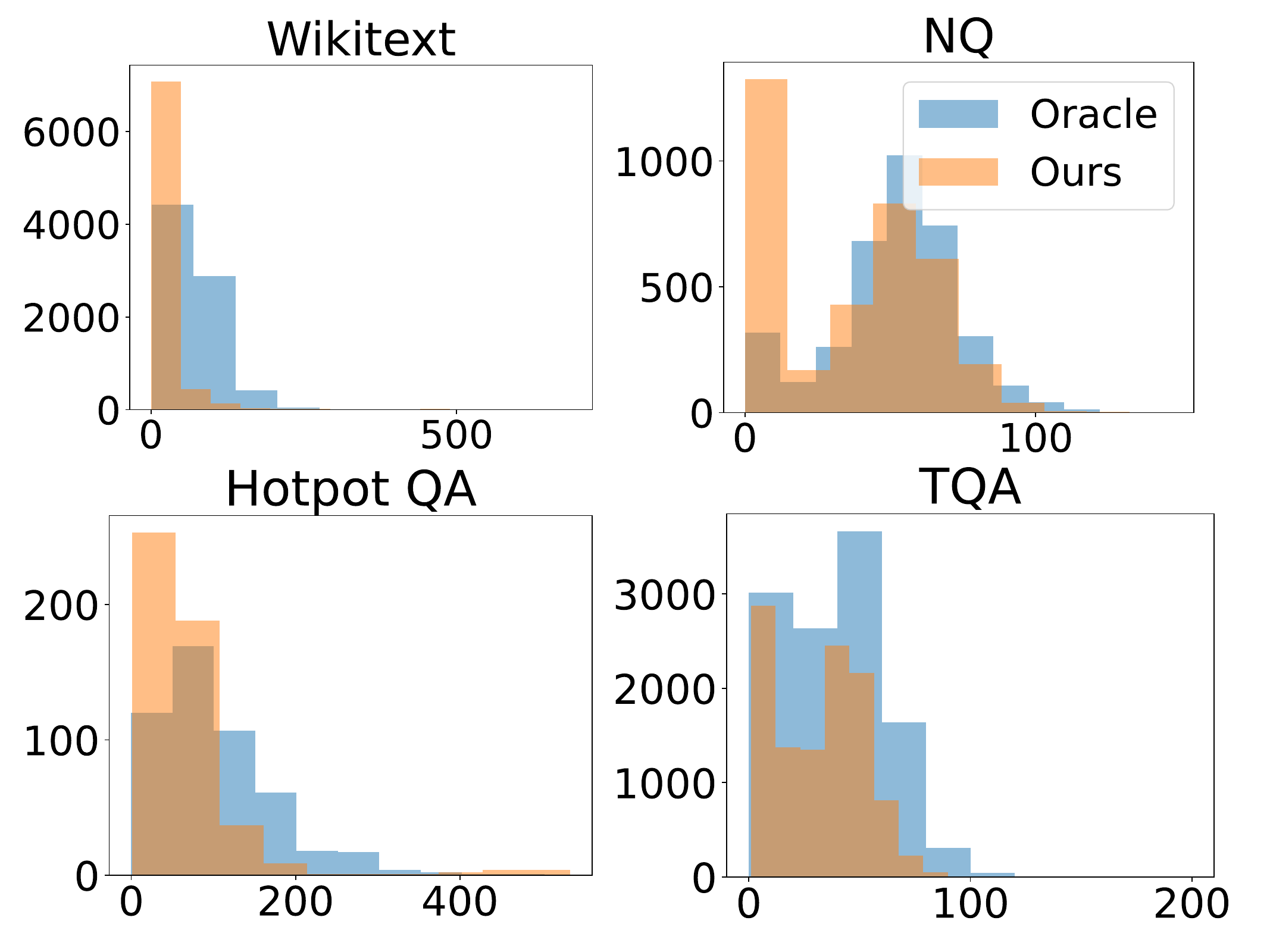}
    \caption{Histogram of abstractive summary length (\# tokens) distribution for testing data of NQ,TQA, HotpotQA and Wikitext.}
    \label{fig:length_hist}
\end{figure}

\begin{table*}[ht!]
\caption{Example abstractive and extractive compression on wikitext-103 dev set and NQ.}
\scriptsize
\begin{tabular}{p{2cm}p{12cm}}
\toprule
\textbf{Wikitext-103 Input} & \textbf{Original Top 1 document} \\ 
present in most of the Mediterranean Sea, only missing from the section east of Crete, and along only the north @-@ west coast of the Black Sea &  Sea of Crete" Sea of Crete The Sea of Crete (, ""Kritiko Pelagos"") or Cretan Sea, is a sea, part of the Aegean Sea, located in its Southern extremity. The sea stretches to the North of the island of Crete, East of the islands of Kythera and Antikythera, South of the Cyclades, and West of the Dodecanese islands of Rhodes, Karpathos and Kassos. The bounding sea to the West is the Ionian Sea. To the Northwest is the Myrtoan Sea, a subdivision of the Mediterranean Sea that lies between the Cyclades and Peloponnese. To the East-SE is the rest of the Mediterranean Sea, \\
\midrule
\textbf{Method} & \textbf{Compressed document}\\ \midrule
BoW &  present Mediterranean Sea missing section east Crete along north west coast Black The Kritiko Pelagos Cretan sea part Aegean located Southern extremity stretches North island East islands Kythera Antikythera South Cyclades West Dodecanese Rhodes Karpathos Kassos bounding Ionian To Northwest Myrtoan subdivision lies Peloponnese SE rest \\ 

NE &  Kythera the Aegean Sea the Ionian Sea Crete Southern South Rhodes the Myrtoan Sea Cretan Sea Antikythera Dodecanese Kassos Karpathos West the Black Sea \&  Sea of Crete Peloponnese the Mediterranean Sea Cyclades \\ 

Extractive compression &  To the Northwest is the Myrtoan Sea, a subdivision of the Mediterranean Sea that lies between the Cyclades and Peloponnese.\\ 
\midrule
\textbf{NQ Input} & \textbf{Original Top 5 document} \\ 
 who got the first nobel prize in physics &  receive a diploma, a medal and a document confirming the prize amount. Nobel Prize in Physics The Nobel Prize in Physics () is a yearly award given by the Royal Swedish Academy of Sciences for those who have made the most outstanding contributions for mankind in the field of physics. It is one of the five Nobel Prizes established by the will of Alfred Nobel in 1895 and awarded since 1901; the others being the Nobel Prize in Chemistry, Nobel Prize in Literature, Nobel Peace Prize, and Nobel Prize in Physiology or Medicine. The first Nobel Prize in Physics was \newline science, Ernest Lawrence won the Nobel Prize in Physics in 1939. Lars Onsager won the 1968 Nobel Prize in Chemistry. Norman Borlaug, father of the Green Revolution, won the Nobel Peace Prize in 1970. Christian B. Anfinsen won the Nobel Prize for chemistry in 1972. Ivar Giaever won the Nobel Prize in Physics 1973. Carl Richard Hagen is noted for his work in physics. In engineering, Clayton Jacobson II is credited with the invention of the modern personal watercraft. Ole Singstad was a pioneer of underwater tunnels. Ole Evinrude invented the first outboard motor with practical commercial application, recognizable today  Nobel Prize in Physics The Nobel Prize in Physics () is a yearly award given by the Royal Swedish Academy of Sciences for those who have made the most outstanding contributions for mankind in the field of physics. It is one of the five Nobel Prizes established by the will of Alfred Nobel in 1895 and awarded since 1901; the others being the Nobel Prize in Chemistry, Nobel Prize in Literature, Nobel Peace Prize, and Nobel Prize in Physiology or Medicine. The first Nobel Prize in Physics was awarded to physicist Wilhelm R00f6ntgen in recognition of the extraordinary services he \newline was also awarded the Abel prize. In addition, eight \"normaliens\" have gone on to receive the Nobel Prize in Physics: Claude Cohen-Tannoudji, Pierre-Gilles de Gennes, Albert Fert, Alfred Kastler, Gabriel Lippmann, Louis N \newline 00e9el, Jean Baptiste Perrin and Serge Haroche, while other ENS physicists include such major figures as Paul Langevin, famous for developing Langevin dynamics and the Langevin equation. Alumnus Paul Sabatier won the Nobel Prize in Chemistry. A ranking of universities worldwide based on ratios of alumni to Nobel prize-winners published in 2016 by American scholars Stephen Hsu and Jonathan Wai placed ENS as the first university worldwide, far \newline rendered by the discovery of the remarkable rays (or x-rays). This award is administered by the Nobel Foundation and widely regarded as the most prestigious award that a scientist can receive in physics. It is presented in Stockholm at an annual ceremony on 10 December, the anniversary of Nobel's death. Through 2018, a total of 209 individuals have been awarded the prize. Only three women (1.4\% of laureates) have won the Nobel Prize in Physics: Marie Curie in 1903, Maria Goeppert Mayer in 1963, and Donna Strickland in 2018. Alfred Nobel, in his last will and testament, stated that his \\
\midrule
\textbf{Method} & \textbf{Compressed document}\\ \midrule
T5 & Wilhelm Röntgen received the first Nobel Prize in Physics in recognition of his extraordinary services. It is one of the five Nobel Prizes established by Alfred Nobel in 1895 and awarded since 1901. \\ 
GPT-3.5-turbo &  The first Nobel Prize in Physics was awarded to physicist Wilhelm Röntgen in 1901 for his discovery of the remarkable rays (or x-rays). Since then, 209 individuals have been awarded the prize, with only three women (1.4\% of laureates) having won it.\\
\bottomrule
\end{tabular} \vspace{-0.3em}
\label{tab:compression_example}
\end{table*}

\begin{table*}[ht!]
\caption{Example input to the Flan-UL2 for NQ with in-context examples and retrieved documents.}
\begin{tabular}{p{2cm}p{12cm}}
\toprule
\textbf{Dataset} & \textbf{Prompts} \\ 
\midrule
NQ &     who won a million on deal or no deal
    Answer: Tomorrow Rodriguez
    
    who is the woman washing the car in cool hand luke
    Answer: Joy Harmon
    
    who is the actor that plays ragnar on vikings
    Answer: Travis Fimmel
    
    who said it's better to have loved and lost
    Answer: Alfred , Lord Tennyson
    
    name the first indian woman to be crowned as miss world
    Answer: Reita Faria
    
    \texttt{Retrieved Docs} 
    
    \texttt{Question} 
    
     Answer: \\
\bottomrule
\end{tabular} \vspace{-0.3em}
\label{tab:icl_prompt}
\end{table*}

\begin{table*}[ht!]
\caption{Prompts used to generated summaries from GPT-3.5-turbo. \texttt{query} and \texttt{docs} represent the actual input query and retrieved documents.}
\begin{tabular}{p{2cm}p{12cm}}
\toprule
\textbf{Dataset} & \textbf{Prompts} \\ 
\midrule
NQ & Compress the information in the retrieved documents into a 2-sentence summary that could be used to answer the question: Question: \texttt{query} Retrieved documents: \texttt{docs} Compressed documents:\\ 
TQA & Compress the information in the retrieved documents into a 2-sentence summary that could be used to answer the question: Question: \texttt{query} Retrieved documents: \texttt{docs} Compressed documents:\\ 
HotpotQA & Source documents: \texttt{docs} Question: \texttt{query} Generate a reasoning chain to answer the question:\\ 
Wikitext & Generate the next two sentences of the given query using the information from the provided documents. \texttt{\textbackslash n}Source Documents: \texttt{docs \textbackslash n}Query: {}\texttt{query \textbackslash n} \\ 
Wikitext & Select sentences from the retrieved docs that are most likely be in the next sentence.\texttt{\textbackslash n}Source Documents: \texttt{docs \textbackslash n}Query: {}\texttt{query\textbackslash n} \\ 
Wikitext & Generate the next one sentence of the given query using the information from the provided documents\texttt{\textbackslash n}Source Documents: \texttt{docs \textbackslash n}Query: {}\texttt{query \textbackslash n} \\
Wikitext & Summarize the information from the provided documents\texttt{\textbackslash n}Source Documents: \texttt{docs \textbackslash n}Query: {}\texttt{query\textbackslash n} \\
\bottomrule
\end{tabular} \vspace{-0.3em}
\label{tab:wikitext_prompt}
\end{table*}

\begin{table*}[ht!]
\caption{Case study of how compressing the retrieved documents helps the model to identify the right answer from NQ dev set.}
\scriptsize
\begin{tabular}{p{2cm}p{9cm}p{3cm}}
\toprule
\multicolumn{2}{l}{\textit{Question: host of the late show who was once a correspondent for the daily show.}} &  \textit{Gold answer: Stephen Colbert} \\
\toprule
\textbf{Type} & \textbf{In-context documents} & \textbf{Predicted Answers} \\ 
\midrule
None &  & Chelsea Handler	 \\ \hline
Top 5 & by Conan O\'Brien, in 2009. Leno explained that he did not want to see a repeat of the hard feelings and controversy that occurred when he was given the show over David Letterman following Carson\'s retirement in 1992. O\'Brien\'s last "Late Night" episode was taped on February 20, 2009. Former Saturday Night Live alum Jimmy Fallon took over as host of "Late Night with Jimmy Fallon" on March 2, 2009. The Colbert Report that aired four days a week on Comedy Central from October 17, 2005, was hosted by Stephen Colbert, one of the regulars on Comedy Central\'s The Daily
season as host began with a notable interview with former British prime minister Tony Blair. The live interview occurred the night before a book signing at Eason's which attracted international attention when Blair was pelted with shoes and eggs and successfully evaded an attempted citizen's arrest on charges of war crimes. On 1 February 2013, Pat Kenny returned to host that night's edition when Tubridy's father died. In 2015, Tubridy's tone and choice of questions when interviewing Anti-Austerity Alliance TD Paul Murphy in relation to the campaign against the implementation of a water tax was much criticised. Opponents of the
'Michigan, interviewing Eminem. Colbert has been given near-full control of the show, with little interference from CBS management in regard to format. Colbert brought most of his staff from "The Colbert Report" with him to "The Late Show", as well as outsiders such as Brian Stack, who is best known for his work on Conan O\'Brien\'s programs, and Jon Stewart, former host of Colbert\'s previous sister program "The Daily Show", who is credited as executive producer. Colbert no longer uses the character he had portrayed on "The Colbert Report", jokingly remarking to Jeb Bush that "I used to play a 
Show" has had three regular hosts: Gay Byrne, Pat Kenny and Ryan Tubridy. Frank Hall deputised for Byrne for one season in the 1960s. There have been at least four occasions on which another presenter has hosted the show. The first was when Byrne became unexpectedly and seriously ill. Frequent panelist Ted Bonner presented instead. The second time was towards the end of a show about feminism, when Byrne ushered a young Marian Finucane into his seat to present the remainder of the show. On another occasion, radio broadcaster and former news reader Andy O\'Mahony replaced Byrne for an interview
popular acclaim. Colbert would host the program until he was chosen to replace David Letterman as host of CBS\'s "Late Show" in 2015. Ed Helms, a former correspondent from 2002 to 2006, also starred on NBC\'s "The Office" and was a main character in the 2009 hit "The Hangover". After filling in as host during Stewart\'s two-month absence in the summer of 2013, John Oliver went on to host his own show on HBO, "Last Week Tonight with John Oliver". In 2016, former correspondent Samantha Bee launched her own late-night talk show "Full Frontal with Samantha Bee". Bee\'s husband Jason & Samantha Bee \\ 
\midrule 
GPT-3.5-turbo & Former Daily Show correspondent Stephen Colbert was chosen to replace David Letterman as host of CBS's "Late Show" in 2015, while Ed Helms, a former correspondent from 2002 to 2006, also starred on NBC's "The Office" and John Oliver, who filled in as host during Jon Stewart's absence in 2013, went on to host his own show on HBO. \\ \midrule
T5 (ours) & Stephen Colbert was a former correspondent for The Daily Show and later became host of CBS's "Late Show" in 2015. He has since brought most of his staff from "The Colbert Report" with him to "The Late Show", with little interference from CBS management in regard to format. & Stephen Colbert \\
\bottomrule
\end{tabular} \vspace{-0.3em}
\label{tab:case_study}
\end{table*}

\begin{table}
\caption{Open-domain QA results on LLaMA-13B. We report the results of oracle compressions with Flan-UL2, the base model for the compressors, (\textit{Oracle w/ FLAN}) and the oracle compression results for LLaMA-13B.}
\small
\begin{center}
\begin{tabular}{@{}lccccccccc@{}}
\toprule
 & \multicolumn{3}{c}{\textbf{NQ}} & \multicolumn{3}{c}{\textbf{TQA}} & \multicolumn{3}{c}{\textbf{HotpotQA}} \\
 \textbf{In-context evidence} & \textbf{\# tok} & \textbf{EM} & \textbf{F1} & \textbf{\# tok} & \textbf{EM} & \textbf{F1} & \textbf{\# tok} & \textbf{EM} & \textbf{F1} \\
\midrule
        - & 0 &  30.89 & 40.73 & 0 & 65.00 & 71.18 & 0 &  24.20 &  34.50\\ 
\midrule
 \multicolumn{10}{l}{\textit{\textbf{RALM without compression}}} \\
       Top 1 document & 132 &  33.35 &  43.13 & 136 &  66.62 &  73.10 & 138 &  34.40 & 44.17  \\
       Top 5 documents & 660 & 37.04 & 47.60 & 667 & 70.61 & 77.51 & 684 & 37.00 & 47.11 \\
\midrule
 \multicolumn{10}{l}{\textit{\textbf{Phrase / token level compression}}} \\
       Top 5 documents (BoW) & 450 & 33.05 & 43.36 & 259 & 66.59 &  73.40 & 255 &  30.00 & 39.13 \\
       Top 5 documents (NE) & 338 & 34.60 &  44.91 & 128 &  65.88 & 72.59 & 157 & 29.20 & 37.93 \\
 \midrule
 \multicolumn{10}{l}{\textit{\textbf{Extractive compression of top 5 documents}}} \\
        \textit{Oracle}  &  31 &  56.62 &  68.89 & 31 & 84.61 & 80.46 & 69 & 42.20 & 51.34 \\ 
        \textit{Oracle (w/ FLAN)} & 34 & 40.89 & 50.06  & 32 & 68.52 &  74.96 & 70 & 35.20 & 45.13 \\ 
         Random & 32 & 30.33 & 39.85 & 31 & 62.80 & 69.25 &  61 & 27.40 & 36.27\\ 
         Contriever  & 36 & 32.52 & 42.01 & 40 & 65.88 & 72.44 & 78 & 34.60 & 43.99 \\
         Ours (init. w/ Contriever) & 37 &  34.38 &  44.15 &  38 & 65.28 & 71.85 & 75 & 33.20 & 42.88\\ 
 \midrule 
 \multicolumn{10}{l}{\textit{\textbf{Abstractive compression of top 5 documents}}} \\ 
         \textit{Oracle} & 50 & 45.60 & 84.87 & 38 & 74.37 & 79.83 & 98 & 41.40 & 51.54 \\ 
         \textit{Oracle (w/ FLAN)} & 51 & 38.98 & 49.40 & 37 & 69.86 & 76.46 &  102 & 35.40 & 46.17 \\
         T5 & 10 & 33.38 & 43.54 & 7 & 63.18 & 70.92 & 7 & 30.40 & 40.60\\ 
         Ours (init. w/ T5) & 36 & 36.32 & 46.10 & 32 & 66.27 & 73.12 & 81 & 30.80 & 40.61 \\ 
        
\bottomrule
\end{tabular} 
\end{center}\vspace{-0.3em}
\label{tab:llama2_results}
\end{table}

\begin{table*}[ht!]
\caption{Example summaries and their manual analysis labels. See Table \ref{tab:example_summaries_2} for more example.}
\scriptsize
\begin{tabular}{p{1cm}p{1cm}p{9cm}p{3cm}}
\toprule
\textbf{Dataset} & \textbf{Model} & \textbf{Query, Passages and Summary} & \textbf{Evaluation} \\
\midrule
NQ & Ours & \textbf{Question:} when will miraculous ladybug season 2 episode 12 come out 

\textbf{Passages:} 2016 on TVNZ's TV2. In Japan, Disney Channel streamed the episode "Stormy Weather" through its mobile application on 1 July 2018, before the official premiere on 23 July in the same year. The second season premiere is scheduled for a global launch around September–November 2017 in Europe, At a panel at San Diego Comic-Con 2017, it was announced that the second season would have its North American release on Netflix in December 2017, with 13 episodes to be released. KidsClick will start airing season 2 of this show in the US starting 30 August 2018, marking the first time that
Korea on 1 September 2015 on EBS1. In the United States, the series debuted on Nickelodeon on 6 December. In the United Kingdom and Ireland, the show premiered on 30 January 2016 on Disney Channel. A Christmas special was released in 2016 and the second season premiered in French on TF1 and in English on Disney Channel UK in 2017. Netflix acquired the U.S. video-on-demand streaming rights and further seasons are in production. Set in modern-day Paris, the series focuses on teenagers Marinette Dupain-Cheng and Adrien Agreste. When evil arises, Marinette transforms into her superhero persona Ladybug, while Adrien transforms
rights in Europe, Eastern Europe and Scandinavia, and free-to-air rights in Spain, Germany, Russia and Turkey. South Korea was the first country to premiere "Ladybug", with girl group Fiestar to sing its translated theme song. It aired on 1 September 2015 on EBS1, and ran for 13 episodes until November 2015, with repeats through February 2016, and its second half of the season airing from 1 March 2016. SK Broadband, having participated in the production, provided the episodes on video on demand exclusively to subscribers of their IPTV platform B TV, about a half-hour following the South Korean broadcast of
Season 2 of this show airing on American over-the-air television. A third season is also in development. On 22 January 2018, Zag posted on Instagram that the crew was working on a fourth and fifth season. On 31 August premiered a new series with Miraculous: Tales of Ladybug \& Cat Noir which this time is in chibi style. On 5 September 2018, Skydance Media has acquired the live-action film and live-action television rights. Kimberly Cooper, a blog writer who has contributed to news media such as "The Huffington Post", wrote that the show has inspired teens and adults to create
Channel on 1 November 2016. In the United Kingdom and Ireland, the show premiered on 30 January 2016 on Disney Channel The first series was also made available to Netflix subscribers in the UK. Among the free-to-air terrestrial television broadcasters in the Republic of Ireland, Raidió Teilifís Éireann premiered the show in 2018 on RTÉ2's teenage programming block, TRTÉ. In Australia, the Australian Broadcasting Corporation released the show on its ABC iview streaming service on 1 January 2016, and premiered the show on its linear channel ABC3 on 22 March 2016. In New Zealand, the show premiered on 27 April
\textbf{Summary:} Miraculous Ladybug season 2 episode 12 is scheduled for release in December 2017 on Netflix in North America. The show premiered in South Korea on 1 September 2015 and has since been aired in various countries, including in the United States, the UK, and New Zealand.

&  Not faithful, Comprehensive \\ 
\midrule
TQA & Ours & \textbf{Question:} In which novel does Flora Poste move to live with her relatives, the Starkadders, in Sussex?

\textbf{Passages:} to be a writer, decides that the only way for her to live whilst researching her writing is to stay with relatives. Her city-based relatives show no interest, so she sends letters to her country relatives. There are a few responses, most of them unsuitable, but one is intriguing. Flora decides to stay for a while with the Starkadder family on their rundown farm. The Starkadders are an assortment of rustic, uncouth, and truly eccentric characters, each of whom has a hurdle (be it physical, emotional, or spiritual) to overcome before reaching his or her potential. Flora quickly realises that,
relatives at the isolated Cold Comfort Farm in the fictional village of Howling in Sussex. The inhabitants of the farm – Aunt Ada Doom, the Starkadders, and their extended family and workers – feel obliged to take her in to atone for an unspecified wrong once done to her father. As is typical in a certain genre of romantic 19th-century and early 20th-century literature, each of the farm's inhabitants has some long-festering emotional problem caused by ignorance, hatred, or fear, and the farm is badly run. Flora, being a level-headed, urban woman in the dandy tradition, determines that she must
Have Always Been Starkadders at Cold Comfort Farm"", set several years later and based on ""Conference at Cold Comfort Farm"", when Flora is married with several children, was broadcast. In 1995 a television film was produced which was generally well-received, with critics. Janet Maslin in the ""New York Times"" wrote that this screen version ""gets it exactly right"". The film starred Kate Beckinsale as Flora, Joanna Lumley as her friend and mentor Mary Smiling, Rufus Sewell as Seth, Ian McKellen as Amos Starkadder, Eileen Atkins as Judith, Stephen Fry as Mybug, Miriam Margolyes as Mrs. Beetle, and Angela Thorne as
dies suddenly of a heart attack and Lady Place is rented out, with the view that Titus, once grown up, will return to the home and run the business. After twenty years of being a live-in aunt Laura finds herself feeling increasingly stifled both by her obligations to the family and by living in London. When shopping for flowers on the Moscow Road, Laura decides she wishes to move to the Chiltern Hills and, buying a guide book and map to the area, she picks the village of Great Mop as her new home. Against the wishes of her extended
respects to her brother. Agnis is moving to the ancestral family home in Newfoundland, which has been abandoned for 44 years. Realizing that Quoyle is at a total loss through grief, she first offers to stay a few more days and help him through the crisis, and then persuades him to move with her. While struggling to rebuild his life, fix up the derelict house, and care for his daughter, Quoyle meets local resident Wavey Prowse, a widow who has a pre-teen boy with a learning disability. Wavey's son and Quoyle's daughter become friends, while the two adults become friends,

\textbf{Summary: } Flora Poste moves to live with her relatives, the Starkadders, on their rundown farm in Sussex. The Starkadders are eccentric characters with various obstacles to overcome.
& Faithful, Not comprehensive \\ 

\bottomrule
\end{tabular} \vspace{-0.3em}
\label{tab:example_summaries}
\end{table*}

\begin{table*}[ht!]
\caption{Example summaries and their manual analysis labels (continued).}
\scriptsize
\begin{tabular}{p{1cm}p{1cm}p{9cm}p{3cm}}
\toprule
\textbf{Dataset} & \textbf{Model} & \textbf{Query, Passages and Summary} & \textbf{Evaluation} \\
\midrule

HotpotQA & GPT-3.5 & \textbf{Question: } The composer of the music for the ballet "The Seasons"  was the director of what organization from 1905 to 1928?

\textbf{Passages: } The Seasons (ballet) The Seasons (, ""Vremena goda""; also ) is an allegorical ballet in one act, four scenes, by the choreographer Marius Petipa, with music by Alexander Glazunov, his Op. 67. The work was composed in 1899 and first performed by the Imperial Ballet in 1900 in St. Petersburg, Russia. The score for Marius Petipa's ""Les Saisons"" (""The Seasons"") was originally intended to have been composed by the Italian composer and conductor Riccardo Drigo, who was Glazunov's colleague and close friend. Since 1886, Drigo held the posts of director of music and ""chef d’orchestre"" to the Ballet of the
harmonium, guitar and even mandolin). ""The Seasons"" was commenced shortly after the premiere of Tchaikovsky's First Piano Concerto, and continued while he was completing his first ballet, ""Swan Lake"". In 1875, Nikolay Matveyevich Bernard, the editor of the St. Petersburg music magazine ""Nouvellist"", commissioned Tchaikovsky to write 12 short piano pieces, one for each month of the year. Bernard suggested a subtitle for each month's piece. Tchaikovsky accepted the commission and all of Bernard's subtitles, and in the December 1875 edition of the magazine, readers were promised a new Tchaikovsky piece each month throughout 1876. The January and February pieces
The Seasons (Cage) The Seasons is a ballet with music by John Cage and choreography by Merce Cunningham, first performed in 1947. It was Cage's first piece for orchestra and also the first to use what Cage later called the ""gamut"" technique, albeit in an early form. Cage composed the music in early 1947, in the midst of working on ""Sonatas and Interludes"". A piano version was first completed, and an orchestral arrangement followed. Cage dedicated ""The Seasons"" to Lincoln Kirstein. The ballet was premiered on May 17, 1947 by the Ballet Society (by which the work was commissioned) at
a new orchestra version. Morton Gould retained the piano part for many of the pieces and orchestrated the work throughout, recording it with himself at the piano in 1951 for American Columbia. In 1965, Kurt-Heinz Stolze orchestrated a number of the pieces as part of the music for John Cranko's ballet ""Onegin"". More recent orchestral versions have been produced by David Matthews (for symphony orchestra), Peter Breiner (for solo violin and symphony orchestra), and Georgii Cherkin (for solo piano and symphony orchestra). French composer Philippe Sarde arranged the Barcarolle as a main theme for the 1988 movie ""The Bear"". Slava
Cranko was not invited to direct, and Sir John Gielgud was brought in. Prosecuted for homosexual activity, Cranko left the UK for Stuttgart, and in 1961 was appointed director of the Stuttgart Ballet, where he assembled a group of talented performers such as Marcia Haydée, Egon Madsen, Richard Cragun, Birgit Keil and Suzanne Hanke. Among his following choreographies were ""Romeo and Juliet"" by William Shakespeare in 1962, set to music by Prokofiev, ""Onegin"" in 1965, an adaptation of the verse novel ""Eugene Onegin"" by Alexander Pushkin, set to music by Tchaikovsky (mainly ""The Seasons""), orchestrated by Kurt-Heinz Stolze, ""The Taming

\textbf{Summary:} 1. The composer of the music for the ballet ""The Seasons"" is Alexander Glazunov.

2. Alexander Glazunov composed the music for the ballet ""The Seasons"" in 1899.

3. The ballet ""The Seasons"" was first performed by the Imperial Ballet in 1900.

4. From 1905 to 1928, the director of an organization was Marius Petipa.

5. Marius Petipa was the choreographer of the ballet ""The Seasons"".

6. Therefore, the composer of the music for the ballet ""The Seasons"" was the director of the Imperial Ballet from 1905 to 1928.

& Not Faithful, Comprehensive \\ 
\bottomrule
\end{tabular} \vspace{-0.3em}
\label{tab:example_summaries_2}
\end{table*}

\end{document}